\documentclass{article}

\PassOptionsToPackage{numbers}{natbib}
\usepackage[preprint]{neurips_2024}

\usepackage[utf8]{inputenc}
\usepackage[T1]{fontenc}
\usepackage{hyperref}
\usepackage{url}
\usepackage{booktabs}
\usepackage{amsfonts}
\usepackage{amsmath}
\usepackage{amssymb}
\usepackage{mathtools}
\usepackage{nicefrac}
\usepackage{microtype}
\usepackage{xcolor}
\usepackage{graphicx}
\usepackage{enumitem}

\title{Beyond Weighted Summation: Learnable Nonlinear Aggregation
       Functions for Robust Artificial Neurons}

\author{
Berke Deniz Bozyigit \\
Independent Researcher \\
\texttt{berkebzt@gmail.com} \\
}

\begin{document}

\maketitle

\begin{abstract}
Weighted summation has remained the default input aggregation mechanism
in artificial neurons since the earliest neural network models.
While computationally efficient, this design implicitly behaves like a
mean-based estimator and is therefore sensitive to noisy or extreme inputs.
This paper investigates whether replacing fixed linear aggregation with
learnable nonlinear alternatives can improve neural network robustness
without sacrificing trainability.
Two differentiable aggregation mechanisms are introduced: an \emph{F-Mean}
neuron based on a learnable power-weighted aggregation rule, and a
\emph{Gaussian Support} neuron based on distance-aware affinity weighting.
To preserve the optimisation stability of standard neurons, hybrid neurons
are proposed that interpolate between linear and nonlinear aggregation
through a learnable blending parameter.
Evaluated in multilayer perceptrons and convolutional neural networks on
CIFAR-10 and a noisy CIFAR-10 variant with additive Gaussian corruption,
hybrid neurons consistently improve robustness under noise while F-Mean
hybrids also yield modest gains on clean data.
The three-way hybrid achieves robustness scores of up to $0.991$ compared
to $0.890$ for the standard baseline, and learned parameters converge
consistently to sub-linear aggregation ($p \approx 0.43$--$0.50$) and
high novelty utilisation ($\alpha \approx 0.69$--$0.79$).
These findings suggest that neuron-level aggregation is a meaningful and
underexplored design dimension for building more noise-tolerant neural
networks.
\end{abstract}

\section{Introduction}

Artificial neurons typically compute a weighted sum of their inputs
followed by a nonlinear activation.
Despite decades of progress in architectures, optimisation algorithms,
and scaling, this aggregation rule has remained nearly unchanged.
Yet weighted summation is not a neutral choice: it behaves like a scaled
arithmetic mean over transformed inputs, inheriting the mean's sensitivity
to outliers and corrupted observations.

This raises a natural question: should every neuron in a network be
constrained to aggregate information in the same linear way, regardless
of task, architecture, or data quality?
In noisy settings, mean-like aggregation may over-react to spurious
activations, whereas more robust or context-sensitive aggregation could
suppress unreliable evidence.
The standard neuron model computes
\begin{equation}
  y = \varphi\!\left(\sum_{i=1}^{n} w_i x_i + b\right),
  \label{eq:standard}
\end{equation}
where $\mathbf{x}\in\mathbb{R}^n$ are inputs, $\mathbf{w}\in\mathbb{R}^n$
are learnable weights, $b$ is a bias, and $\varphi$ is a nonlinear
activation.
The inner summation is, up to a scale factor and the bias, the weighted
arithmetic mean of the inputs, which is the unique minimiser of the
least-squares objective $J(\mu)=\sum_i(w_i x_i - \mu)^2$.
For over 70 years, this formulation has remained the default building
block of virtually every neural network architecture.

This paper revisits the aggregation stage of the artificial neuron.
Two differentiable, learnable alternatives are proposed that retain
compatibility with gradient-based training while broadening the family
of computations available at the neuron level.
The focus is on two mechanisms: a learnable power-based aggregation that
smoothly adjusts how strongly large responses are emphasised, and a
Gaussian support aggregation that weights inputs according to agreement
in the transformed feature space.
To reduce optimisation risk, each nonlinear aggregator is combined with
the standard linear path through a learnable hybrid blend.

The contributions of this paper are as follows.
\begin{itemize}[leftmargin=1.2em]
  \item Two differentiable alternatives to fixed sum aggregation are
        formulated: F-Mean aggregation and Gaussian Support aggregation.
  \item Hybrid neurons are introduced that learn how much to rely on
        standard linear aggregation versus nonlinear alternatives.
  \item These designs are evaluated in both MLP and CNN settings on clean
        and noisy CIFAR-10.
  \item Hybrid neurons are shown to consistently improve robustness to
        corruption, with F-Mean hybrids additionally offering small
        clean-data gains.
  \item Learned parameter convergence is analysed, revealing interpretable
        sub-linear aggregation strategies that emerge without explicit
        regularisation.
\end{itemize}

\section{Related Work}

Most neural architectures inherit the classical neuron design in which
aggregation is fixed to weighted summation.
Prior work has explored robust estimators such as median-like aggregation,
generalised mean or $p$-norm style operators, and adaptive weighting
mechanisms related to attention or kernel methods.
These approaches suggest that aggregation matters, but they often
emphasise either robustness or flexibility in isolation, or incur
non-trivial computational costs.

\paragraph{Generalised neuron models.}
Yadav et al.~\cite{yadav2006generalised} proposed generalised-mean
neurons that replace summation with a power mean of fixed exponent.
Baldeschi et al.~\cite{baldeschi2019double} introduced a double-weight
neuron definition that separates magnitude and direction in the
aggregation step.
Fan et al.~\cite{fan2023neuroai} surveyed neuronal diversity in
biological systems and argued for incorporating diverse computational
primitives into artificial networks.
The present work extends these ideas by making the aggregation exponent
itself a trainable parameter and combining it with distance-based
weighting within a hybrid framework.

\paragraph{Learnable aggregation in graph neural networks.}
Pellegrini et al.~\cite{pellegrini2021learning} introduced Learnable
Aggregation Functions (LAFs), a parametric family that approximates
diverse statistical behaviours including sums, means, and higher-order
moments.
Corso et al.~\cite{corso2020principal} proposed Principal Neighbourhood
Aggregation, which combines multiple aggregators to improve the
expressive power of graph neural networks.
These methods target graph-structured data and typically rely on
implicit-layer solvers that introduce non-trivial computational overhead.
The architectures in this paper target standard feed-forward neurons
and use closed-form aggregations that avoid iterative sub-problems.

\paragraph{Robust aggregation.}
Geisler et al.~\cite{geisler2020reliable} introduced the Soft Medoid,
a differentiable robust aggregation function for graph neural networks
that achieves a high breakdown point against adversarial perturbations.
Pillutla et al.~\cite{pillutla2022robust} applied the geometric median
to federated learning, improving resilience to corrupted client updates.
Both of these works select a fixed robust estimator.
The present approach instead learns which aggregation behaviour is most
appropriate for each layer, allowing the network to adapt to the
distribution of its own activations.

\paragraph{Learnable pooling and attention.}
Gülçehre et al.~\cite{gulcehre2014learned} studied learned-norm pooling
in deep feedforward and recurrent networks.
Terziyan et al.~\cite{terziyan2022hyper} explored hyper-flexible CNNs
based on Lehmer and power means, demonstrating that generalised pooling
can improve performance with modest computational overhead.
The Gaussian Support neuron draws additional inspiration from attention
mechanisms~\cite{niu2021review} and Radial Basis Function networks,
extending distance-based weighting to individual neurons in standard
fully-connected and convolutional architectures.

Rather than replacing the standard neuron outright, linear aggregation
is treated as a strong baseline, and learnable nonlinear extensions are
asked to improve performance, especially under noisy inputs.

\section{Method}

\subsection{Design Criteria}

Any candidate aggregation function must satisfy three properties to be
practical in gradient-based training.
It must include at least one \emph{learnable parameter} that modifies
its aggregation behaviour during training.
It must be \emph{fully differentiable} with respect to all inputs and
parameters to enable backpropagation.
And the \emph{arithmetic mean must be a special case}, so that the
neuron can be initialised to standard behaviour and explore alternative
strategies gradually.

\subsection{F-Mean Neuron}
\label{sec:fmean}

The F-Mean neuron replaces fixed summation with a power-weighted mean.
Let $z_i = w_i x_i$ denote the scaled inputs, and define
softplus-transformed values
\begin{equation}
  z^+_i = \ln(1 + e^{z_i}),
\end{equation}
which are guaranteed to be positive.
The power-normalised aggregation weights are
\begin{equation}
  \omega^{(p)}_i = \frac{(z^+_i)^p}{\sum_j (z^+_j)^p + \varepsilon},
  \qquad \varepsilon = 10^{-8},
  \label{eq:fmean_weights}
\end{equation}
and the aggregation output is
\begin{equation}
  A_{\text{F-Mean}}(\mathbf{x}) = \sum_i \omega^{(p)}_i \cdot z_i.
  \label{eq:fmean_agg}
\end{equation}
The parameter $p \in \mathbb{R}$ is learnable and assigned per output
unit.
When $p = 1$, the weights become proportional to $z^+_i$, recovering
approximately uniform weighting similar to standard linear aggregation.
As $p \to 0$, weights approach uniform values, producing harmonic-like
averaging.
As $p \to \infty$, weight concentrates on the maximum-valued input,
producing max-like behaviour.
Smaller values of $p$ induce sub-linear behaviour that downweights
large activations; this is the regime that networks consistently
prefer in practice.

The softplus transformation ensures $z^+_i > 0$, preventing undefined
gradients at zero crossings, and the epsilon term in
Equation~\ref{eq:fmean_weights} prevents division by zero.
Parameters are initialised at $p = 1$ to provide a stable starting
point that matches the standard neuron.

\subsection{Gaussian Support Neuron}
\label{sec:gaussian}

The Gaussian Support neuron weights each input by its aggregate
similarity to all other inputs in the transformed feature space.
The pairwise affinity between inputs $i$ and $j$ is
\begin{equation}
  \mathrm{Aff}(i,j) = \exp\!\left(
    -\frac{\|z_i - z_j\|^2}{2\sigma^2}
  \right),
  \label{eq:affinity}
\end{equation}
where $\sigma > 0$ is a learnable width parameter stored as $\log\sigma$
for unconstrained optimisation.
Support weights are obtained by summing affinities and normalising:
\begin{equation}
  \alpha_i = \frac{\sum_j \mathrm{Aff}(i,j)}{\sum_k \sum_j \mathrm{Aff}(k,j)}.
  \label{eq:support_weights}
\end{equation}
The aggregation is then $A_{\text{Gauss}}(\mathbf{x}) = \sum_i \alpha_i z_i$.

Large $\sigma$ yields broad, nearly uniform weighting; smaller $\sigma$
favours locally consistent responses.
Pairwise distance computation has $O(n^2)$ complexity, which is managed
through a dimensionality-reduction projection layer applied before the
Gaussian neuron.

\subsection{Hybrid Neurons}
\label{sec:hybrid}

Pure nonlinear aggregation may be harder to optimise than the standard
linear neuron.
Hybrid neurons are therefore defined that interpolate between linear and
nonlinear paths.
The two-way hybrid combines a novel aggregation $A_{\text{novel}}$
with standard linear aggregation via a learnable scalar:
\begin{equation}
  y = \phi\!\left(\tilde\alpha \cdot A_{\text{novel}}(\mathbf{x})
    + (1 - \tilde\alpha) \cdot A_{\text{linear}}(\mathbf{x}) + b\right),
  \quad \tilde\alpha = \sigma(\alpha_{\text{raw}}) \in (0, 1).
  \label{eq:twoway}
\end{equation}
The three-way hybrid combines all three aggregations through
softmax-normalised coefficients:
\begin{equation}
  y = \phi\!\left(\sum_{k=1}^{3} \tilde\alpha_k \cdot A_k(\mathbf{x}) + b\right),
  \quad \tilde{\boldsymbol{\alpha}} = \mathrm{softmax}(\boldsymbol{\alpha}_{\text{raw}}),
  \label{eq:threeway}
\end{equation}
where $A_1$, $A_2$, $A_3$ denote linear, F-Mean, and Gaussian
aggregations respectively.

Initialising $\alpha_{\text{raw}} = 0$ sets $\tilde\alpha = 0.5$,
providing equal initial weight to both pathways.
If novel aggregation is unhelpful, the network can reduce $\tilde\alpha$
and revert toward linear behaviour.
If it is beneficial, $\tilde\alpha$ grows during training.
The blending parameter also acts as a natural regulariser, dampening
gradient instability from the novel aggregation path.

\section{Experimental Setup}

\subsection{Architectures}

Experiments are conducted in two model families.

\paragraph{MLP.}
A projection layer first reduces input dimensionality ($3072 \to 128$),
followed by a hidden layer using the chosen hybrid neuron and a standard
linear classifier.

\paragraph{CNN.}
Standard convolutional feature extraction is preserved, and hybrid
neurons are introduced in the classifier head after flattening
($8192 \to 256$ via projection).
This isolates the effect of alternative aggregation while keeping
convolutional processing unchanged.

\subsection{Datasets}

Experiments use CIFAR-10, comprising $50{,}000$ training and $10{,}000$
test images across 10 classes.
To assess robustness, a noisy variant is constructed by injecting
additive Gaussian noise with standard deviation $\sigma_{\text{noise}}
= 0.15$ at test time.

\subsection{Training Details}

Four aggregation settings are compared: standard linear, F-Mean hybrid,
Gaussian hybrid, and three-way hybrid, across two architectures and two
data conditions, yielding 16 experimental configurations in total.
Standard weight parameters use learning rate $10^{-3}$; novel parameters
($\alpha$, $p$, $\log\sigma$) use $10^{-2}$ to encourage exploration.
Gradient clipping with maximum norm $1.0$, a ReduceLROnPlateau scheduler,
and early stopping with patience 10 are applied throughout.
Numerical stability is supported by softplus transformations,
log-space parameterisation for $\sigma$, and $\varepsilon = 10^{-8}$
regularisation.

\subsection{Evaluation}

Clean accuracy, noisy accuracy, and a robustness score are reported:
\begin{equation}
  \rho = \frac{\text{Accuracy}_{\text{noisy}}}{\text{Accuracy}_{\text{clean}}}.
\end{equation}
Learned aggregation parameters are additionally analysed to understand
whether models converge toward linear or nonlinear behaviour.

\section{Results}

\subsection{MLP Performance}

Table~\ref{tab:mlp} summarises results for the MLP setting.
All hybrid variants improve clean accuracy over the baseline, with
F-Mean and three-way hybrids reaching approximately $55.2\%$ compared
to $52.3\%$ for the baseline.
The three-way hybrid achieves the highest robustness score of $0.991$,
showing the smallest performance drop under noise.
This suggests that combining multiple aggregation behaviours is
especially useful in low-capacity models where robustness is otherwise
difficult to obtain.

\begin{table}[t]
  \centering
  \caption{MLP results on clean and noisy CIFAR-10.}
  \label{tab:mlp}
  \begin{tabular}{lccc}
    \toprule
    Model & Clean (\%) & Noisy (\%) & $\rho$ \\
    \midrule
    Baseline        & 52.30 & 51.45 & 0.984 \\
    F-Mean Hybrid   & \textbf{55.17} & 53.56 & 0.971 \\
    Gaussian Hybrid & 54.30 & 53.30 & 0.982 \\
    Three-way Hybrid & 55.21 & \textbf{54.72} & \textbf{0.991} \\
    \bottomrule
  \end{tabular}
\end{table}

\subsection{CNN Performance}

Table~\ref{tab:cnn} presents the CNN results.
The F-Mean hybrid achieves the best clean and noisy accuracy, while
the three-way hybrid attains the highest robustness ratio.
Clean-data gains are modest but consistent across training runs,
indicating a real rather than incidental improvement.
All hybrid CNN variants outperform the standard baseline on the
robustness metric, indicating that nonlinear aggregation confers
benefits even when strong convolutional feature extraction is already
present.
The transition from MLP to CNN yields approximately $32$ percentage
points of absolute accuracy gain across all variants, confirming that
hybrid neurons integrate successfully with convolutional architectures.

\begin{table}[t]
  \centering
  \caption{CNN results on clean and noisy CIFAR-10.}
  \label{tab:cnn}
  \begin{tabular}{lccc}
    \toprule
    Model & Clean (\%) & Noisy (\%) & $\rho$ \\
    \midrule
    Standard CNN    & 87.33 & 77.73 & 0.890 \\
    F-Mean CNN      & \textbf{87.61} & \textbf{78.41} & 0.895 \\
    Gaussian CNN    & 86.60 & 77.33 & 0.893 \\
    Three-way CNN   & 86.37 & 77.52 & \textbf{0.898} \\
    \bottomrule
  \end{tabular}
\end{table}

\subsection{Learned Aggregation Behaviour}

The learned parameters reveal consistent trends across all architectures.
For the F-Mean path, the power parameter converges to sub-linear values
in all settings: $p \approx 0.476$ (MLP F-Mean), $0.430$ (CNN F-Mean),
$0.495$ (three-way MLP), and $0.445$ (three-way CNN).
These values indicate a systematic preference for suppressing extreme
activations, a behaviour that was not enforced by regularisation but
emerged purely from gradient-based optimisation.

\begin{table}[t]
  \centering
  \caption{Converged novel parameters across architectures.}
  \label{tab:params}
  \begin{tabular}{lccc}
    \toprule
    Architecture & $\alpha$ & $p$ & $\sigma$ \\
    \midrule
    MLP F-Mean      & 0.762 & 0.476 & -- \\
    CNN F-Mean      & 0.786 & 0.430 & -- \\
    MLP Gaussian    & 0.689 & --    & 4.68 \\
    CNN Gaussian    & 0.785 & --    & 7.30 \\
    Three-way MLP   & 0.70--0.79 & 0.495 & 5.08 \\
    Three-way CNN   & 0.70--0.79 & 0.445 & 8.50 \\
    \bottomrule
  \end{tabular}
\end{table}

The blending coefficient $\alpha$ settles between $0.69$ and $0.79$
across architectures, indicating that models rely substantially on
novel aggregation while retaining a non-trivial linear component.
The similarity of these values across MLP and CNN settings suggests
they represent fundamental optimal blending ratios rather than
architecture-specific artefacts.

The Gaussian width $\sigma$ converges to moderate values in all cases.
CNN architectures converge to larger $\sigma$ values than MLPs,
which is consistent with richer feature representations available after
convolutional processing.
Moderate $\sigma$ produces neither highly localised nor fully global
aggregation, suggesting that useful behaviour lies between these
two extremes.

\section{Discussion}

The results support two main conclusions.
First, neuron-level aggregation is not merely an implementation detail:
replacing fixed summation with learnable alternatives can improve both
performance and robustness.
Second, hybridisation is critical.
Rather than forcing the network to commit to a fully novel neuron, the
blend with standard linear aggregation stabilises optimisation and acts
as a safety mechanism.

The strongest clean-data results come from the F-Mean hybrid, especially
in CNNs, suggesting that sub-linear power-based weighting is a practical
way to improve on mean-like aggregation.
The three-way hybrid is the most robust under corruption, indicating
that complementary aggregation mechanisms can better absorb noisy or
inconsistent evidence.
Three mechanisms contribute to this robustness.
Sub-linear F-Mean aggregation ($p < 1$) compresses large values
relative to small ones, reducing the influence of inputs that are
unusually large due to noise.
Gaussian affinities assign lower weight to inputs that deviate from
their neighbours, providing a continuous analogue of median-like
consensus.
And the $\alpha$ blending parameter allows the network to fall back
toward linear aggregation during unstable phases of training.

The improvements on clean CIFAR-10 are small in absolute terms.
This is expected: the dataset is relatively clean and low-resolution,
and standard CNNs already extract strong features from it.
Alternative aggregation is most valuable where standard neurons face
distribution shift, input noise, or corrupted observations.

\subsection{Computational Overhead}

The F-Mean neuron adds minimal overhead relative to a standard linear
layer, as the additional operations are element-wise or involve small
summations.
The Gaussian Support neuron has $O(n^2)$ complexity due to pairwise
distance computation.
This is managed here with a projection layer, but further optimisation
through sparse attention or approximate nearest-neighbour methods
remains a direction for future work.

\section{Limitations and Future Work}

This study is limited to CIFAR-10 and one controlled noisy variant,
so broader validation is still needed.
The Gaussian support mechanism introduces additional computational cost
due to pairwise interactions, and the present work focuses on
neuron-level modifications in MLPs and CNN classifier heads; other
architectural settings may expose different trade-offs.

Future work should extend evaluation to larger benchmarks such as
ImageNet, medical imaging tasks, and NLP problems.
Comparison against other robust aggregation mechanisms, including
trimmed means and M-estimators, would clarify where each approach
offers the greatest advantage.
Combining learnable aggregation with transformers and attention-based
architectures is a particularly promising direction, given that
aggregation already plays a central role in attention computation.
Theoretical analysis of convergence guarantees and formal robustness
bounds would also strengthen the foundations of this approach.

\section{Conclusion}

This paper revisited a longstanding assumption in neural network design:
that neuron inputs should always be aggregated by weighted summation.
Two learnable nonlinear alternatives were proposed, F-Mean and Gaussian
Support aggregation, and embedded in hybrid neurons that preserve the
optimisation strengths of standard layers.
Across MLP and CNN experiments on CIFAR-10 and noisy CIFAR-10, hybrid
neurons consistently improved robustness, while F-Mean hybrids also
delivered small but repeatable gains on clean data.
Learned parameters converge to interpretable sub-linear strategies
without explicit regularisation, providing evidence that networks
discover more robust aggregation behaviours autonomously when given the
freedom to do so.
These findings suggest that input aggregation is an underexplored and
promising axis of neural architecture design.

\bibliographystyle{abbrvnat}

\end{document}